\documentclass[lettersize,journal]{IEEEtran}
\usepackage{amsmath,amsfonts}
\usepackage{algorithmic}
\usepackage{algorithm}
\usepackage{array}
\usepackage[caption=false,font=normalsize,labelfont=sf,textfont=sf]{subfig}
\usepackage{textcomp}
\usepackage{stfloats}
\usepackage{url}
\usepackage{verbatim}
\usepackage{graphicx}
\usepackage{cite}
\usepackage{mathrsfs}
\usepackage{amsthm,amsmath}
\usepackage{amssymb}
\usepackage{balance}
\usepackage {color}
\newcommand{\ie}{\emph{i.e., }}
\newcommand{\eg}{\emph{e.g., }}
\newcommand{\etal}{\emph{et al.}}

\begin{document}

\title{Sentiment-enhanced Graph-based Sarcasm Explanation in Dialogue}
\author{
	\IEEEauthorblockN{
		Kun Ouyang, 
		Liqiang Jing,~\IEEEmembership{Student Member,~IEEE,}
		Xuemeng Song,~\IEEEmembership{Senior Member,~IEEE,} \\
   Meng Liu,~\IEEEmembership{Member,~IEEE,}
		Yupeng Hu,~\IEEEmembership{Member,~IEEE,}
		Liqiang Nie,~\IEEEmembership{Senior Member,~IEEE}}
\thanks{This work is supported by the National Natural Science Foundation of China, No.:62376137, No.:62376140, No.:62276155, and No.:U23A20315; the Shandong Provincial Natural Science Foundation, No.:ZR2022YQ59; the Science and Technology Innovation Program for Distinguished Young Scholars of Shandong Province Higher Education Institutions, No.:2023KJ128, and the Special Fund for Taishan Scholar Project of Shandong Province; Shenzhen College Stability Support Plan (Grant No. GXWD20220817144428005). (Corresponding author: Xuemeng Song.)}
\thanks{Kun Ouyang and Xuemeng Song are
with the School of Computer Science and Technology, Shandong University, Qingdao 266237, China (e-mail: kunouyang10@gmail.com, sxmustc@gmail.com).}

\thanks{Liqiang Jing is with the Department of Computer Science, University of Texas at Dallas, USA (e-mail: jingliqiang6@gmail.com).}

\thanks{Meng Liu is with the School of Computer Science and Technology, Shandong Jianzhu University, Jinan 250101, China (e-mail:
mengliu.sdu@gmail.com).}

\thanks{Yupeng Hu is with the School of Software, Shandong University, Jinan 250101, China (e-mail: huyupeng@sdu.edu.cn).}
\thanks{Liqiang Nie is with the School of Computer Science and Technology, Harbin Institute of Technology (Shenzhen), Shenzhen 518055, China (e-mail: nieliqiang@gmail.com).}
}

% \markboth{Journal of \LaTeX\ Class Files,~Vol.~14, No.~8, August~2021}

% \IEEEpubid{0000--0000/00\$00.00~\copyright~2025 IEEE}
% \IEEEpubidadjcol
\maketitle
\begin{abstract}
% up to 200 words                 
Sarcasm Explanation in Dialogue (SED) is a new yet challenging task, which aims to generate a natural language explanation for the given sarcastic dialogue that involves multiple modalities (\ie utterance, video, and audio). Although existing studies have achieved great success based on the generative pretrained language model BART, they overlook exploiting the sentiments residing in the utterance, video and audio, which play important roles in reflecting sarcasm that essentially involves subtle sentiment contrasts. Nevertheless, it is non-trivial to incorporate sentiments for boosting SED performance, due to three main challenges: 1) diverse effects of utterance tokens on sentiments; 2) gap between video-audio sentiment signals and the embedding space of BART; and 3) various relations among utterances, utterance sentiments, and video-audio sentiments. To tackle these challenges, we propose a novel sEntiment-enhanceD Graph-based multimodal sarcasm Explanation framework, named EDGE. In particular, we first propose a lexicon-guided utterance sentiment inference module, where a heuristic utterance sentiment refinement strategy is devised. We then develop a module named Joint Cross Attention-based Sentiment Inference (JCA-SI) by extending the multimodal sentiment analysis model JCA to derive the joint sentiment label for each video-audio clip. 
Thereafter, we devise a context-sentiment graph to comprehensively model the semantic relations among the utterances, utterance sentiments, and video-audio sentiments, to facilitate sarcasm explanation generation. Extensive experiments on the publicly released dataset WITS verify the superiority of our model over cutting-edge methods.
\end{abstract}


\section{Introduction}
\begin{figure}
\centering
  \includegraphics[scale=0.52]{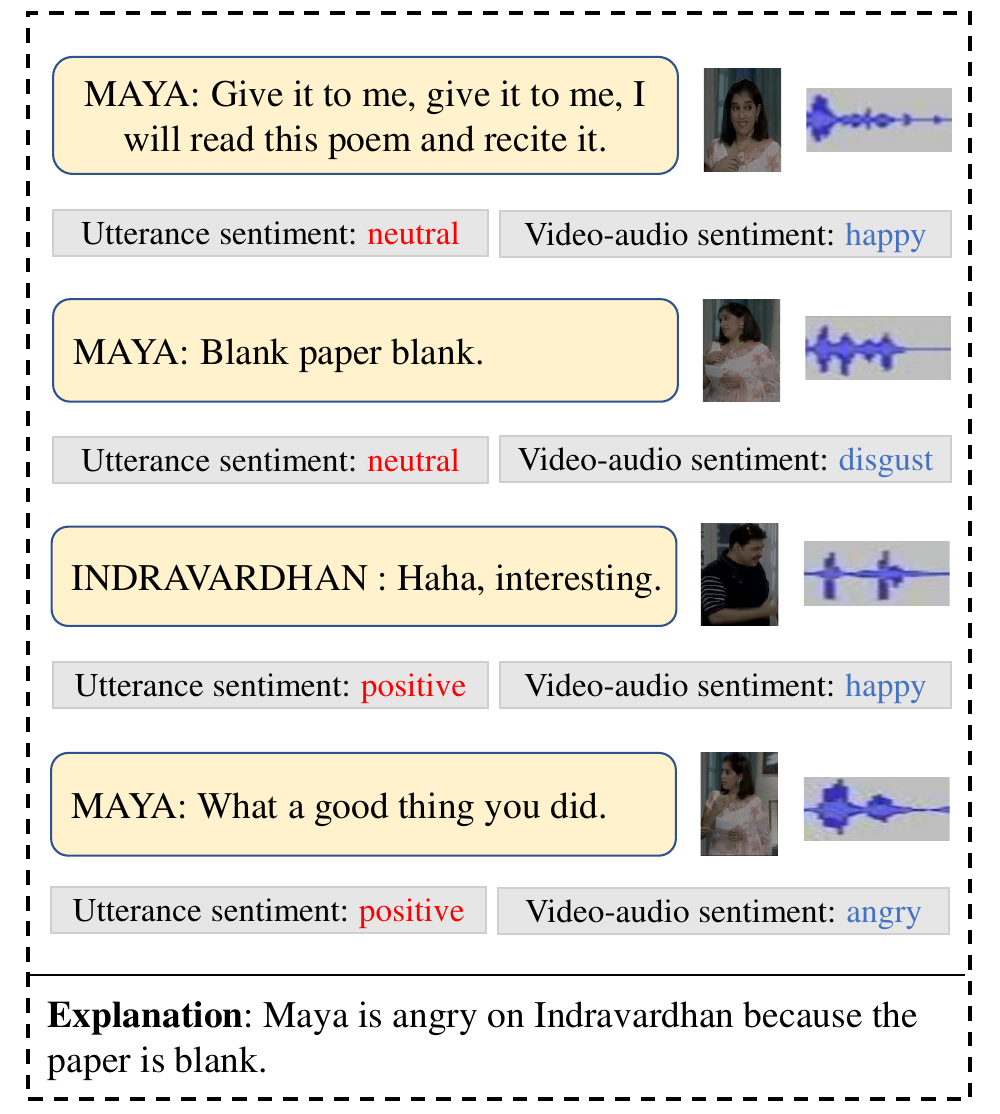}
    % \vspace{-0.5cm}
    \caption{A sample of the sarcasm explanation in dialogue from the WITS dataset~\cite{DBLP:conf/acl/KumarKA022} and the corresponding sentiments. 
    }
     \vspace{-0.3cm}
    \label{fig:intro}
\end{figure}
\IEEEPARstart{T}{he} use of sarcasm in people's daily communication is very common, which is an important method to express people's sentiments or opinions in a contrary manner. 
Therefore, sarcasm explanation is important for understanding people's sentiments~(\eg positive and negative) or opinions conveyed in their daily expressions. 
Due to its great practical value, many researchers~\cite{DBLP:conf/acl/ChakrabartyGMP20,DBLP:conf/aaai/Desai0A22,DBLP:conf/acl/JingSOJN23,DBLP:conf/acl/KumarKA022} have made efforts to sarcasm explanation. For example, Chakrabarty~\etal~\cite{DBLP:conf/acl/ChakrabartyGMP20} employed a retrieve and edit framework, which retrieves factual knowledge and leverages it to edit the input text, thereby generating the sarcasm explanation. Although previous studies on sarcasm explanation have attained impressive results, they focus on investigating the sarcasm explanation for pure textual input. Recently, noticing the rapid development of multimedia and the essential role of video and audio content in conveying sarcasm, Kumar~\etal~\cite{DBLP:conf/acl/KumarKA022} proposed a new Sarcasm Explanation in Dialogue (SED) task. As shown in Fig.~\ref{fig:intro}, SED aims at generating a natural language explanation for a given multimodal sarcastic dialogue that contains the utterance, video, and audio modalities. Existing work~\cite{DBLP:conf/acl/KumarKA022,DBLP:conf/acl/LewisLGGMLSZ20} on SED focus on designing various multimodal fusion methods to effectively inject the video and audio modalities into the generative \mbox{pretrained} language model BART~\cite{DBLP:conf/acl/LewisLGGMLSZ20} for sarcasm explanation generation.

Despite their promising performance, they only consider the content of utterances, video, and audio, but overlook the sentiment information contained in the dialogue.  
In fact, in the context of SED, the sarcastic semantics can be reflected by the inconsistency between the sentiments delivered by utterances and those conveyed by corresponding video-audio clips~\cite{DBLP:conf/lrec/RayMNB22}. 
\textcolor{black}{Fig.~\ref{fig:intro} shows a sample from WITS~\cite{DBLP:conf/acl/KumarKA022} dataset, consisting of four utterances, where the sentiment of each utterance and that of the corresponding video-audio clip are also provided.
As can be seen, for this dialogue, the sarcasm is especially expressed by the last utterance ``What a good thing you did''. 
By referring the provided sentiment labels, we can learn that compared to the former three utterances, the utterance sentiment (\ie ``positive'') of the last utterance is apparently more inconsistent with its video-audio sentiment (\ie ``angry'').
This suggests that the sentiment inconsistency may be a potential indicator of the sarcastic semantics.}
Therefore, in this work, we aim to exploit the sentiments involved in the utterance, video, and audio of the dialogue to assist sarcastic semantic understanding and hence boost the SED performance. Similar to previous work, we also adopt BART as the model backbone because of its strong generation ability.

However, it is non-trivial to enhance SED by exploiting the sentiment information due to the following challenges: 
\textbf{C1: Diverse effects of utterance tokens on sentiments.} 
There are various types of tokens in the utterance, such as turning tokens (\eg ``but''), negating tokens (\eg ``not''), intensity tokens (\eg ``very''), and sentiment tokens (\eg ``happy''), which have diverse contributions to the sentiments of the utterance. Therefore, how to analyze the various effects of these tokens on the utterance sentiments is a vital challenge. 
 \textbf{C2: Gap between video-audio sentiment signals and the embedding space of BART.}  
The sentiment signals delivered by the video and audio modalities, like facial expressions and voice tones, do not match the semantic space of BART well, since BART is pretrained purely on the textual corpus. Therefore, how to effectively inject sentiment information into BART is an important challenge.
 \textbf{C3: Various semantic relations among utterances, utterance sentiments, and video-audio sentiments.} 
There are rich semantic relations among utterances, utterance sentiments, and video-audio sentiments (\eg the semantic association among tokens in utterance and the sentiment inconsistency between the utterance sentiment and its corresponding video-audio sentiment), which can be important cues for sarcasm explanation~\cite{DBLP:conf/lrec/RayMNB22}. How to model these various relations to improve sarcasm explanation generation is also a crucial challenge.

\begin{figure*}
    \centering
    \includegraphics[scale=0.52]{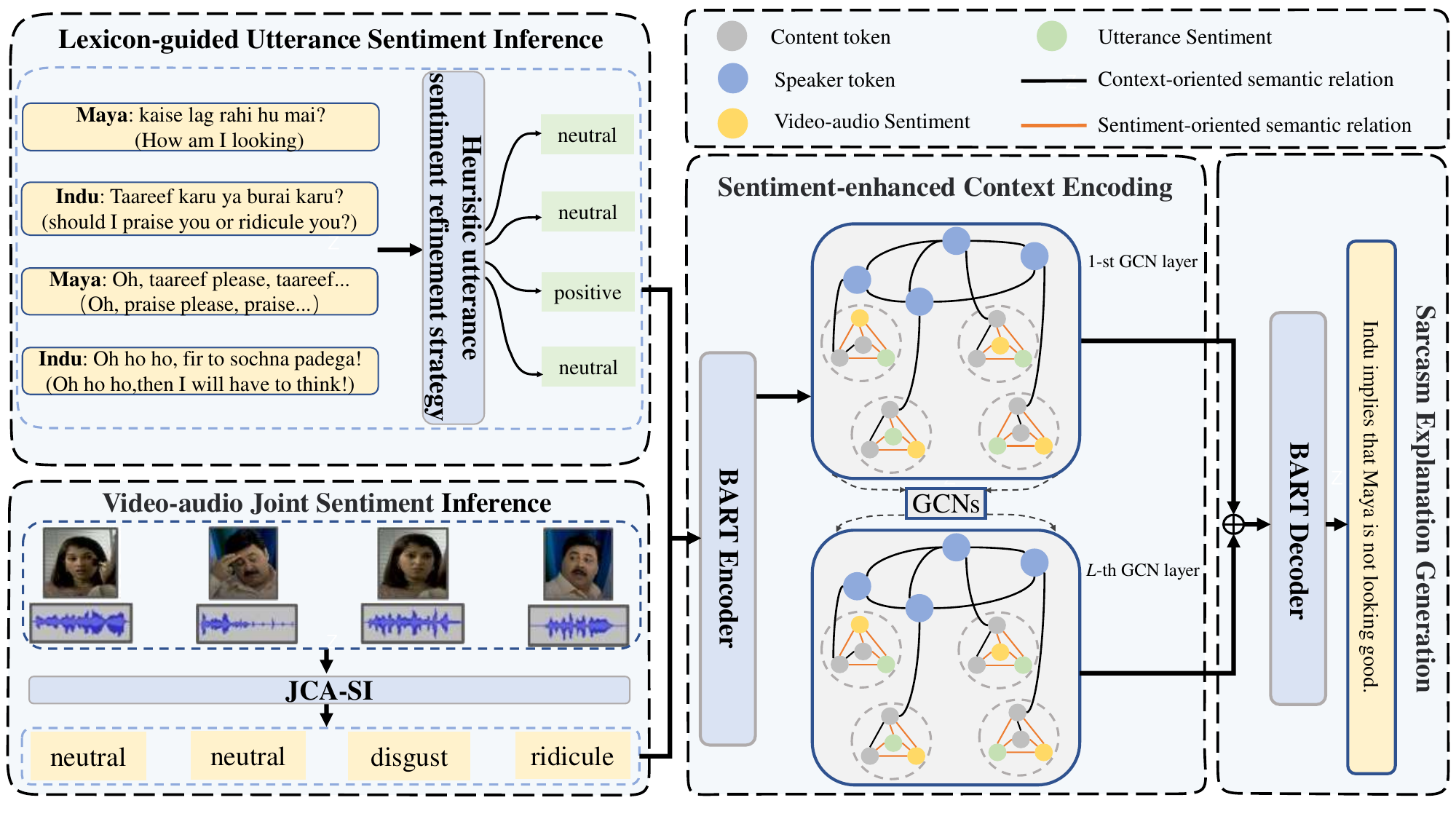}
    \caption{Illustration of the proposed EDGE, which contains four components.
    }
    \label{fig:framework}
\end{figure*}

To address the challenges mentioned above, we propose a novel sEntiment-enhanceD Graph-based multimodal sarcasm Explanation framework, EDGE for short, with BART as the backbone. 
Specifically, EDGE consists of four components: lexicon-guided utterance sentiment inference, video-audio joint sentiment inference, sentiment-enhanced context encoding, and sarcasm explanation generation, as shown in Fig.~\ref{fig:framework}. 
In the first module, we devise a heuristic utterance sentiment refinement strategy to accurately infer the utterance sentiments based on BabelSenticNet~\cite{DBLP:conf/ssci/VilaresPSC18}, which can analyze the various effects of different tokens on the utterance sentiments.
In the second module, we infer the joint sentiment of the video and audio modalities to assist the sarcastic semantic understanding. To make the sentiment information match the semantic space of BART, 
we devise a module named Joint Cross Attention-based Sentiment Inference (JCA-SI) based on the existing multimodal (\ie video and audio) sentiment analysis model JCA~\cite{DBLP:conf/cvpr/PraveenMUAZDPKB22}. Different from the original JCA, our JCA-SI predicts meaningful sentiment labels (\eg ``angry'', ``disgust'', and ``excited'') rather than its original valence and arousal scores to facilitate sentiment understanding of BART.
In the third module, we adopt Graph Convolutional Networks~(GCNs)~\cite{DBLP:conf/iclr/KipfW17} to fulfill the sarcasm comprehension. In particular, we construct a context-sentiment graph to comprehensively model the semantic relations among the utterances, utterance sentiments, and video-audio sentiments, where both context-oriented and sentiment-oriented semantic relations are mined.
In the last module, we adopt the BART decoder to generate the sarcasm explanation. 
We conduct extensive experiments on the public SED dataset and the experimental results show the superiority of our method over existing methods. 
Our contributions can be concluded as follows.
\begin{itemize}
    \item We propose a novel sEntiment-enhanceD Graph-based  multimodal sarcasm Explanation framework, where both utterance sentiments and video-audio sentiments are exploited for boosting the sarcasm understanding. 
    \item We propose a heuristic utterance sentiment refinement strategy that can analyze the various effects of these tokens of the utterance on the sentiments based on BabelSenticNet. 
    \item We propose a context-sentiment graph, which is able to comprehensively capture the semantic relations among utterances, utterance sentiments, and video-audio sentiments. As a byproduct, we release our code and parameters\footnote{\url{https://github.com/OuyangKun10/EDGE}.} to facilitate the research community.
\end{itemize}
\section{Related Work}
\textbf{Sarcasm Detection.}  
Early studies~\cite{DBLP:journals/access/BouaziziO16,DBLP:conf/emnlp/FelboMSRL17} on sarcasm detection mainly utilized hand-crafted features, such as punctuation marks, POS tags, emojis, and lexicons, to detect the sarcastic intention. Later, with the advancement of deep learning methodologies, some researchers turned to neural network architectures for sarcasm detection~\cite{DBLP:conf/acl/SuTHL18,DBLP:conf/coling/BabanejadDAP20}. 
Although these efforts have made promising progress in text-based sarcasm detection, they overlook the fact that multimodal information has been popping up all over the internet. 
In the bimodal setting, sarcasm detection with multimodal posts containing the image and caption was first proposed by Schifanella~\etal~\cite{DBLP:conf/mm/SchifanellaJTC16}, and this work introduces a framework that fuses the textual and visual information with Convolutional Neural Networks~\cite{DBLP:conf/iccv/MaLSL15} to detect sarcasm. Thereafter, researchers~\cite{DBLP:conf/mm/Pentland05,DBLP:conf/aaai/QiaoJSCZN23,DBLP:journals/corr/abs-2312-10493} explored more advanced network architecture for multimodal information fusion to improve multimodal sarcasm detection, such as Graph Neural Networks~(GCNs)~\cite{DBLP:conf/iclr/KipfW17} and Transformer~\cite{DBLP:conf/nips/VaswaniSPUJGKP17}. 
Apart from the multimodal posts, researchers also noticed that sarcasm is commonly used in the dialogue.  
In the dialogue setting, Castro~\etal~\cite{DBLP:conf/acl/CastroHPZMP19} created a multimodal, multispeaker dataset named MUStARD, which is considered the benchmark for multimodal sarcasm detection. To tackle this task, Hasan~\etal~\cite{DBLP:conf/aaai/HasanLR0MMH21} proposed a humor knowledge-enriched transformer model, which achieved state-of-the-art performance on this dataset. Nevertheless, these efforts can only recognize the sarcasm in a dialogue, but cannot explain the underlying sarcastic connotation of the dialogue and capture its true essence, which is also important for various applications~\cite{DBLP:conf/aaai/Desai0A22,DBLP:conf/acl/KumarKA022}, such as media analysis and conversational systems.
\\ \indent
\textbf{Sarcasm Explanation.} 
Apart from sarcasm detection, a few efforts attempted to conduct the sarcasm explanation, which aims to generate a natural language explanation for the given sarcastic post or dialogue. For example, some work~\cite{DBLP:conf/acl/PeledR17,DBLP:conf/comad/DubeyJB19} resorted to machine translation models to generate non-sarcastic interpretation for sarcastic text, which can help the smart customer service understand users' sarcastic comments and posts on various platforms. 
Notably, these methods only focus on text-based sarcasm explanation generation. 
Therefore, Desai~\etal~\cite{DBLP:conf/aaai/Desai0A22} adopted BART~\cite{DBLP:conf/acl/LewisLGGMLSZ20} with a cross-modal attention mechanism to generate sarcasm explanation for multimodal posts. 
Beyond them, recently, Kumar~\etal~\cite{DBLP:conf/acl/KumarKA022} first proposed the novel task of Sarcasm Explanation in Dialogue (SED) and released a dataset named WITS, which targets at generating a natural language explanation for a given sarcastic dialogue. %, to support the sarcasm explanation in dialogue.  
In addition, Kumar~\etal~\cite{DBLP:conf/acl/KumarKA022,DBLP:conf/aaai/KumarMA023} adopted the generative language model BART as the backbone and incorporated the visual and acoustic features into the context information of the dialogue with the multimodal context-aware attention mechanism to solve the SED task. 
Despite its remarkable performance, this method overlooks the sentiments involved in the dialog which can assist the ironic semantics understanding~\cite{DBLP:conf/lrec/RayMNB22}. 

\section{Methodology}
In this section, we first formulate the task of SED, then detail the four components of our proposed EDGE. 
\subsection{Task Formulation}
Suppose we have a training dataset $\mathcal{D}$ composed of $N_d$ training samples, \ie $\mathcal{D}=\{(T_1, A_1, V_1, Y_1), \cdots, (T_{N_d}, A_{N_d}, V_{N_d}, Y_{N_d})\}$. For each sample $(T,V,A,Y)$, $T=\{u_1, u_2, \cdots, u_{N_u}\}$ is the input text containing $N_u$ utterances, $V$ is the input video, $A$ is the corresponding audio, and $Y=\{y_1, y_2, \cdots y_{N_y}\}$  denotes the target explanation text consisting of $N_y$ tokens. In addition, each utterance $u_j=\{s^j_0,t^j_1,\cdots,t^j_{N_{u_j}-1}\}$ contains $N_{u_j}$ tokens, in which the first token $s^j_0$ denotes the corresponding speaker's name and the other tokens are content tokens. 
Based on these training samples, our target is to learn a model $\mathcal{F}$ that can generate the sarcasm explanation in dialogue based on the given multimodal input as follows,
\begin{equation}
    \hat{Y}= \mathcal{F}(T,V,A|\Theta),
\end{equation}
where $\Theta$ is a set of to-be-learned parameters of the model $\mathcal{F}$. $\hat{Y}$ is the generated explanation text. For simplicity, we temporally omit the subscript $i$ that indexes the training samples.

\subsection{Lexicon-guided Utterance Sentiment Inference}
In this module, we extract the sentiment of each utterance, which plays important role in sarcastic semantic understanding~\cite{DBLP:conf/lrec/RayMNB22}. Specifically, we resort to 
BabelSenticNet~\cite{DBLP:conf/ssci/VilaresPSC18}, a large-scale multi-language sentiment lexicon, to obtain the utterance sentiment. \textcolor{black}{It has been widely used for sentiment analysis in previous work~\cite{DBLP:journals/jksucis/ElfaikN23,SanayaiMeetei2021LowRL}.}
In particular, BabelSenticNet provides polarity values of a set of $100$k common natural language concepts. The polarity value is a floating number between $-1$ and $+1$, which reflects the sentiment of the concept. The higher the number, the more positive the sentiment. To drive the utterance sentiment, we first derive the sentiment of each token in the utterance according to BabelSenticNet. Formally, let ${p}_{k}^j$ denote the derived polarity value of the content token $t_k^j$ in the utterance $u_j$, where $k=1,2,\cdots,N_{u_j}-1$. Notably, for tokens not found in BabelSenticNet, we treat them as neutral tokens and set their polarity values to $0$. 

After getting the polarity values of all tokens, one naive method for deriving the utterance sentiment is directly calculating the sum of polarity values of all tokens.
However, this naive method ignores the following three issues. 1) The turning tokens in the utterance can clearly indicate the following sub-sequence plays the essential effect in determining the utterance sentiment. The sub-sequence stressed by the turning token can determine the utterance sentiment. For example, the sentiment of the utterance ``This dessert tastes delicious, but I hate its high price.'' is determined by the stressed sub-sequence ``I hate its high price''. 2) The negating tokens (\eg ``not'' and ``never'' ) can reverse the sentiment of the following sentiment token (\eg ``happy'' and ``angry''). 3) The intensity tokens may strengthen or weaken the utterance sentiment when they modify the sentiment tokens, \eg ``little'' and ``very''.  

\textcolor{black}{To solve the above three issues, we propose a heuristic utterance sentiment refinement strategy, which works on refining the utterance sentiment by modeling specific impacts of turning tokens, negating tokens and intensity tokens on utterance sentiment.
}

\textcolor{black}{First, turning tokens are identified to select the sub-sequence stressed by them, and the selected sub-sequence is then used to determine the utterance sentiment. In particular, we first derive a common turning token set $\mathcal{S}^r$ according to SentiWordNet\footnote{\url{https://github.com/aesuli/SentiWordNet}.}, a widely used lexical resource for sentiment analysis~\cite{DBLP:conf/lrec/BaccianellaES10}.} Then for each utterance $u_j$, we identify its turning token based on the common turning token set $\mathcal{S}^t$. Next, we only adopt the stressed sub-sequence $u^s_j$\footnote{For the selected sub-sequence that still contains turning tokens, we continue this process until there is no turning token in the selected part, to choose the sub-sequence that contributes most to the utterance sentiment.}, which is positioned either before or after the turning token based on the emphatic order indicated in $\mathcal{S}^t$, for the following utterance sentiment inference.

\textcolor{black}{Second, negating tokens are considered to reverse the polarity of the sentiment tokens.} In particular, 
for each sentiment token in the utterance, we check whether the token ahead it is a negating token.  If it is, we reverse the original polarity of the sentiment token as follows,
\begin{equation}
 \hat{p}_k^j=
 \left\{
 \begin{aligned}
-p_k^j & , \\\
p_k^j & ,
\end{aligned}
 \right.
  \begin{aligned}
&if~t^j_{k-1}\in\mathcal{S}^n,\\\
&otherwise,
\label{eq:p_hat1}
 \end{aligned}
\end{equation}
where $\hat{p}_k^j$ is the refined polarity value, $\mathcal{S}^n$ is the negating token set defined according to Sentiwordnet. 

\textcolor{black}{Third, intensity tokens are used for modifying the utterance sentiment intensity by scaling the polarity accordingly with a scaling factor defined in SentiwordNet~\cite{DBLP:conf/lrec/BaccianellaES10}}. To be specific,
for each sentiment token in the utterance, we check whether the token ahead it is an intensity token. 
If it is, we utilize the sentiment scaling factor $\alpha \in (0,2)$ which is a floating number provided by SentiWordNet, to refine the value of the polarity $\hat{p}_k^j$ of the sentiment token. 
Formally, we have
\begin{equation}
 \hat{p}_k^j=
 \left\{
 \begin{aligned}
\alpha \times \hat{p}_k^j & , \\\
\hat{p}_k^j & ,
\end{aligned}
 \right.
  \begin{aligned}
&if~t^j_{k-1}\in\mathcal{S}^i,\\\
&otherwise,
\label{eq:p_hat2}
\end{aligned}
\end{equation}
where $\mathcal{S}^i$ is the intensity token set defined according to SentiWordNet.

\begin{figure*}[h]
    \centering
    \includegraphics[scale=0.35]{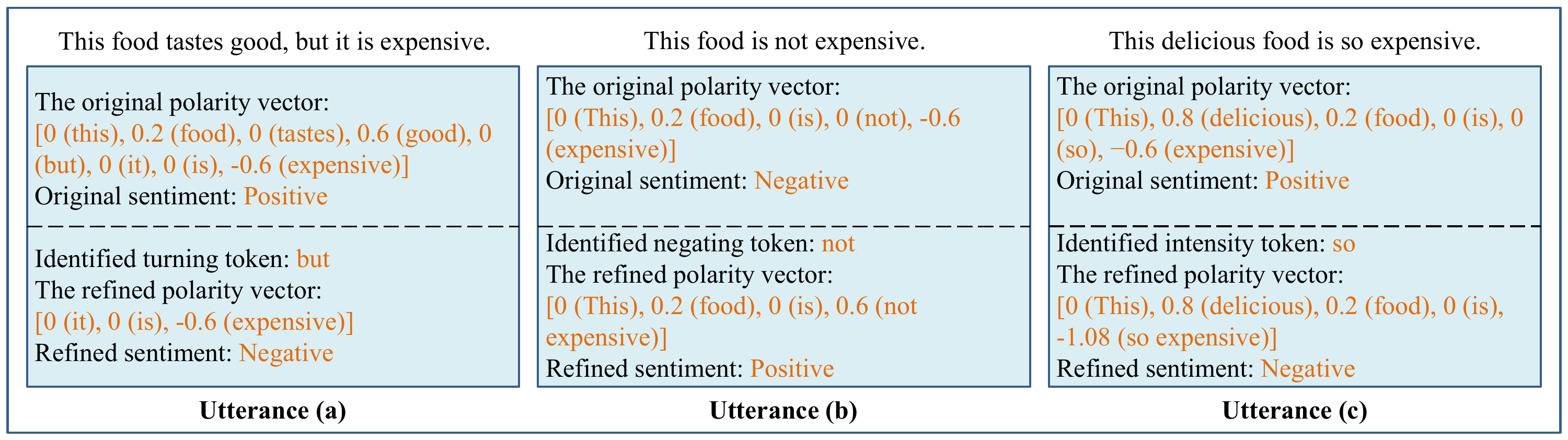}
    \caption{The utterance sentiment inference process for three example utterances. And we compare the refined sentiments with the original sentiments.}
    \label{fig:sentiment}
\end{figure*}

Based on the above process, we can obtain the refined polarity vector $\boldsymbol{\hat{p}_j}=[\hat{p}_1^j,\hat{p}_2^j,\cdots,\hat{p}_{N_{u_j}}^j]$, where $N_{u_j}$ denotes the number of tokens in $u_{j}$. Finally, we can sum  the elements of the refined polarity vector $\boldsymbol{\hat{p}_j}$ to identify the sentiment of the utterance $u_j$ as follows,
\begin{equation}
e^T_j=
 \left\{
 \begin{aligned}
&0, \\\
&1,\\\
&2,
\end{aligned}
 \right.
  \begin{aligned}
&if~\operatorname{sum}(\boldsymbol {\hat{p}_j})\textgreater 0,\\\
&if~\operatorname{sum}(\boldsymbol {\hat{p}_j})= 0,\\\
&if~\operatorname{sum}(\boldsymbol {\hat{p}_j})\textless 0,
\label{eq:emotion}
\end{aligned}
\end{equation}
where $0$, $1$, and $2$ refer to positive, neutral, and negative, respectively, as the sentiment label of the input utterance. $sum(\boldsymbol{\hat{p}_j})$ is the sum of the elements in $\boldsymbol{\hat{p}_j}$.
Then for the input text $T=\{u_1,  u_2, \cdots, u_{N_u}\}$, we can obtain the corresponding sentiment labels, denoted as $E^T=\{e_1^T,  e_2^T, \cdots, e_{N_u}^T\}$, where $N_u$ is the total number of utterances. Fig.~\ref{fig:sentiment} shows three examples for utterance sentiment inference.
\subsection{Video-audio Joint Sentiment Inference}\label{sec:Sentiment,V-A}
It has been proven that the jointly utilization of the sentiment conveyed in both video and audio can improve the efficacy of sentiment inference~\cite{DBLP:journals/tmm/NieRNZ21,9797846,10214099,DBLP:journals/tmm/WangLWTHG23}. Therefore,
we propose to jointly extract the video-audio sentiment to promote SED.  

In detail, we introduce a variant of a Joint Cross-Attention Model (JCA)~\cite{DBLP:conf/cvpr/PraveenMUAZDPKB22}, named Joint Cross Attention-based Sentiment Inference, JCA-SI for short. Notably, JCA is a multimodal sentiment analysis model, which utilizes an advanced attention mechanism to recognize the sentiment information involved in the video and audio~\cite{DBLP:conf/lrec/RayMNB22}. Although it shows great performance in the task of multimodal sentiment analysis~\cite{DBLP:conf/cvpr/PraveenMUAZDPKB22,DBLP:journals/tmm/NieCRSL22}, it can only predict two types of sentiment value (\ie valence and arousal), which are float number ranging from -$1$ to $1$.
If we directly utilize JCA to conduct video-audio joint sentiment inference, the predicted sentiment value may not match the semantic space of BART. The reason is that BART cannot capture the sentiment information involved in the sentiment value as it does not learn the meaning of the sentiment value during the pre-training phase. Therefore, we devise a variant named JCA-SI. Specifically, we add a multi-layer perceptron to conduct sentiment classification after obtaining the feature representation via JCA in order to convert the sentiment value into sentiment label.
In fact, video-audio sentiment changes for different utterances in the long video and audio of the whole dialogue as it contains multiple utterances. It is crucial to align the video, audio and utterance so that the video-audio sentiments and the utterance sentiments are one-to-one correspondence. This alignment facilitates the extraction of inconsistency between the video-audio sentiment and utterance sentiment. 
Therefore, we segment the video $V$ of the whole dialogue into $N_u$ video clips $\{v_1, v_2, \cdots, v_{N_u}\}$ based on temporal annotations provided by WITS, each clip $v_j$ is corresponding to an utterance $u_j$. Similarly, we conduct the same process for the audio $A$ of the whole dialogue, and obtain $N_u$ audio clips $\{a_1, a_2, \cdots, a_{N_u}\}$. 

Next, we feed video clips $\{v_1, v_2, \cdots, v_{N_u}\}$ and audio clips $\{a_1, a_2, \cdots, a_{N_u}\}$ to visual and acoustic feature extraction modules in the JCA model, respectively.
For the video modality, we resort to I3D~\cite{DBLP:conf/cvpr/CarreiraZ17} to extract the features of each video clip $v_j$. 
For the audio modality, 
we feed the audio clip $a_j$ to Resnet~18~\cite{DBLP:conf/cvpr/HeZRS16} to get the audio feature. 
Formally, we have
\begin{equation}
 \left\{
 \begin{aligned}
& \boldsymbol{X}_v^j=\operatorname{I3D}\left(v_j\right), \\\
&\boldsymbol{X}_a^j=\operatorname{Resnet 18}\left(a_j\right),
\end{aligned} \label{eq:AudioVideo_feat}
 \right.
\end{equation}
where $\boldsymbol {X}_a^j \in \mathbb{R}^{d_a\times N_c}$ and $\boldsymbol{X}_v^j\in \mathbb{R}^{d_v\times N_c}$ represent two feature matrixes extracted from the segmented audio clip $a_j$ and the segmented video clip $v_j$, respectively. $d_a$ and $d_v$ refer to the feature dimension of the audio and video representation, respectively. $N_c$ denotes the resampled clip size of the segmented audio clip $a_j$ and the segmented video clip $v_j$. We then concatenate $\boldsymbol{X}_a^j$ and $\boldsymbol{X}_v^j$ to obtain $\boldsymbol{J}=[\boldsymbol{X}_a^j;\boldsymbol{X}_v^j]\in \mathbb{R}^{d\times N_c}$, where $d=d_a+d_v$.
Next, we feed $\boldsymbol{X}_a^j$, $\boldsymbol{X}_v^j$ and $\boldsymbol{J}$ to the joint cross attention layer~\cite{DBLP:conf/cvpr/PraveenMUAZDPKB22} to calculate the attended visual features $\boldsymbol{\hat{X}}_v^j$ and the attended acoustic features $\boldsymbol{\hat{X}}_a^j$, respectively. 
Mathematically, 
\begin{equation}
 \left\{
 \begin{aligned}
&\boldsymbol{\hat{X}}_v^j=\operatorname{Att}\left(\boldsymbol{X}_v^j,\boldsymbol{J}\right),\\\
&\boldsymbol{\hat{X}}_a^j=\operatorname{Att}\left(\boldsymbol{X}_a^j,\boldsymbol{J}\right),
\end{aligned} \label{eq:AudioVideo_attention}
 \right.
\end{equation}
where $\operatorname{Att}(\cdot)$ denotes the joint cross attention layer. \textcolor{black}{It can be defined as follows:
\begin{equation}
\left\{
\begin{aligned}
     \boldsymbol{C}_m&=\operatorname{tanh}\left(\frac{(\boldsymbol{X}_m^j)^\top\boldsymbol{W}_{om}\boldsymbol{J}}{\sqrt{d}}\right),\\\
    \boldsymbol{H}_m&=\operatorname{ReLu}\left(\boldsymbol{W}_m\boldsymbol{X}_m^j + \boldsymbol{W}_{cm}\boldsymbol{C}_m^\top\right),\\\
    \boldsymbol{\hat{X}}_m^j&=\boldsymbol{W}_{hm}\boldsymbol{X}_m^j + \boldsymbol{X}_m^j,
\end{aligned}
 \right.
\end{equation}
where $\boldsymbol{W}_{om}\in \mathbb{R}^{N_c\times N_c}, \boldsymbol{W}_m\in \mathbb{R}^{s\times N_c}, \boldsymbol{W}_{cm}\in \mathbb{R}^{s\times d} ~\text{and}~\boldsymbol{W}_{hm}\in \mathbb{R}^{s\times N_c}$ represent the learnable weight matrices, $m\in\{a,v\}$. $\boldsymbol{C}_m$ is the joint correlation matrix, while $\boldsymbol{H}_m$ represents the attention maps. $\operatorname{tanh}$ and $\operatorname{ReLu}$ are the activation functions.
}

Finally, we feed the attended visual features $\boldsymbol{\hat{X}}_v$ and the attended acoustic features $\boldsymbol{\hat{X}}_a$ to the sentiment classification network and obtain the corresponding sentiment as follows, 
\begin{equation}
e^{V\text{-}A}_j =\operatorname{MLP}([\boldsymbol{\hat{X}}_v^j;\boldsymbol{\hat{X}}_a^j]),
\end{equation}
where $\operatorname{MLP}(\cdot)$ is a multi-layer perceptron to achieve sentiment classification. It consists of two fully connected layers followed by a $softmax$ activation function to compute the probability distribution of each sentiment, including angry, sad, frustrated, ridicule, disgust, excited, fear, neutral, surprised and happy. \textcolor{black}{$e^{V\text{-}A}_j$ is the video-audio sentiment label for the $j$-$th$ video-audio clip.}
For $V=\{v_1, v_2, \cdots, v_{N_u}\}$ and $A=\{a_1, a_2, \cdots, a_{N_u}\}$, we can obtain a set of video-audio sentiment label $e_i^{V\text{-}A}$ corresponding to each pair $(v_i,a_i)$, \ie $E^{V\text{-}A}=\{e_1^{V\text{-}A},e_2^{V\text{-}A},\cdots,e_{N_u}^{V\text{-}A}\}$.
\begin{figure*} 
    \centering
    \includegraphics[scale=0.51]{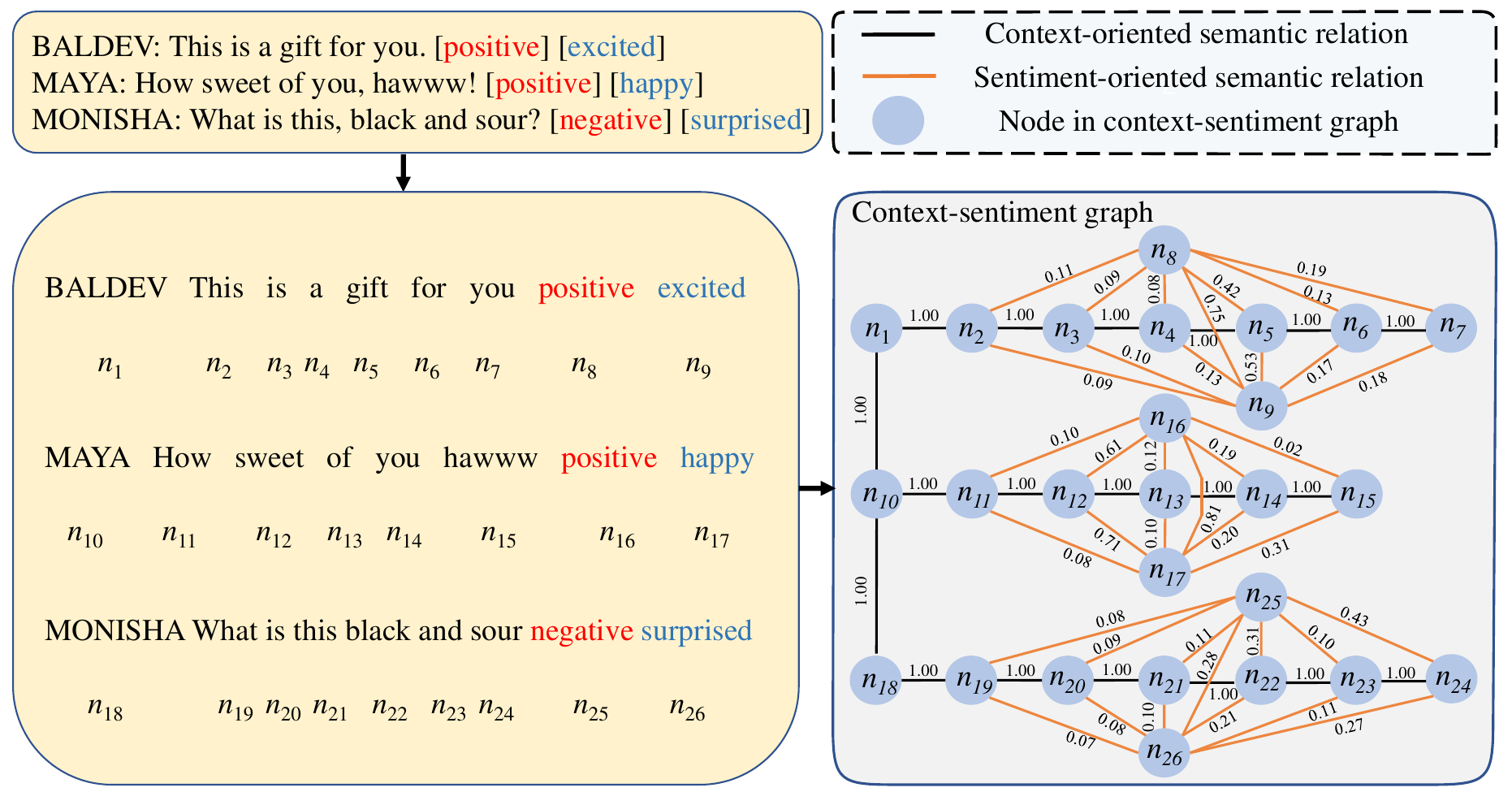}
    \caption{\textcolor{black}{The example of a context-sentiment graph, which is constructed for a dialogue including three utterances. Tokens in red are the utterance sentiments and those in blue are video-audio sentiments. $n_j$ denotes the $j$-$th$ node in the context-sentiment graph.}}
    \label{fig:graph}
\end{figure*}
\subsection{Sentiment-enhanced Context Encoding} 
In this module, we aim to enhance the context encoding with the extracted utterance sentiment labels and video-audio sentiment labels.
To this end,  we resort to the widely used graph neural networks (GCNs)~\cite{DBLP:conf/iclr/KipfW17}, to mine the rich semantic relations among the given utterance sequence, its corresponding utterance sentiment labels, and video-audio sentiment labels. 
Specifically, we first build a novel context-sentiment graph $\mathcal{G}$.

\subsubsection{Nodes Initialization} In particular, the nodes in the context-sentiment graph $\mathcal{G}$ come from three kinds of sources, the given utterances $T$, extracted utterance sentiment labels $E^T$, and extracted video-audio sentiment labels $E^{V\text{-}A}$. All the nodes can be defined as $\{n_1,\cdots,n_N\}=\{T, E^{T}, E^{V\text{-}A}\}$, 
To initialize the nodes, we resort to the BART encoder~\cite{DBLP:conf/acl/LewisLGGMLSZ20}
to extract the features of the utterances, utterance sentiment labels and video-audio sentiment labels.
Specifically, we first concatenate them into a sequence of tokens, denoted as $X=\{T, E^T, E^{V\text{-}A}\}$, and  then feed $X$ into the BART encoder $\mathcal{E}$ as follows,
\begin{equation}
   \mathbf{H}=\mathcal{E}(X),\label{encoding}
\end{equation} 
where $\mathbf{H}=[\mathbf{h}_1, \cdots, \mathbf{h}_N] \in \mathbb{R}^{N \times D}$ is the encoded representation matrix, each column of which corresponds to a token. $N$ is the total number of tokens in $X$. Accordingly, nodes in the context-sentiment graph $\mathcal{G}$ can be initialized by $ \mathbf{H}$, where the $j$-$th$ token node is initialized with  $\mathbf{h}_j$.

\subsubsection{Semantic Relation Construction} 
To promote the context encoding with extracted sentiment labels, we consider two kinds of semantic relations: context-oriented semantic relation and sentiment-oriented semantic relation. The former captures the basic information flow of the given dialog, and the latter enables the injection the sentiment information into the utterance content.

\textbf{Context-oriented Semantic Relation}. 
To capture the information flow of the given context, \ie the utterance sequence in the given dialogue $\{u_1,u_2,\cdots,u_{N_u}\}$, and promote the context understanding, we design three types of context-oriented semantic edges. a) \textit{Speaker-speaker edges}. We connect the same speaker in different utterances with an edge and the adjacent speakers with an edge. \textcolor{black}{b) \textit{Speaker-token edges}. We connect an edge between the speaker node and the first content token node in the utterance to represent the matching relation between the speaker and the utterance.} c) \textit{Token-token edges}. \textcolor{black}{We introduce an edge between each pair of adjacent content token nodes in the utterance to represent the neighboring relations among the tokens of utterance.} The above edges characterize the information flow, and thus weighted by $1$. Formally, we introduce the corresponding adjacency matrix $\mathbf{A}^1$ for representing these edges as follows,
\begin{equation}
 \mathbf{A}^1_{i,j}=\left\{
 \begin{aligned}
&1,\quad if~{D_1}(n_i,n_j), \\\
&0,\quad otherwise,
\end{aligned}\label{eq:graph}
 \right.
\end{equation}
where $N_t$ denotes the total number of tokens in the input text $T$ and $i,j \in [1, N_t]$. $D_1(n_i,n_j)$ denotes that the nodes $n_i$ and  $n_j$ have certain context-oriented semantic relation.

\textbf{Sentiment-oriented Semantic Relation}.
To fully utilize both the utterance sentiment labels and video-audio sentiment labels for promoting the sarcastic semantic understanding, we design the following three types of edges. 
a)  \textit{Utterance sentiment-content edges}. For each utterance sentiment node, we link it with each content token  in the utterance to capture their semantic relations. The rational is to inject the utterance sentiment information into the context of dialogue. b) \textit{Video-audio sentiment-content edges}. Similarly, for each video-audio sentiment node,  we connect it to each content token in the corresponding utterance. c)  \textit{Sentiment-sentiment edges}. We introduce an edge between the utterance sentiment node and the video-audio sentiment node of the same utterance, to excavate the sentiment inconsistency between them.

To adaptively utilize the sentiment information, we introduce a weight for each sentiment-oriented semantic relation. The philosophy is that, given an edge, the higher the semantic/sentiment similarity between two tokens the edge links, the higher edge weight should be assigned.
Formally, we have 
\begin{equation}
    w(n_i,n_j)=\min(1, Sim(t_i,t_j)/\lvert p_i-p_j\rvert), 
\end{equation}
where $t_i$ and $t_j$ denote the corresponding tokens of nodes $n_i$ and $n_j$, respectively. $Sim(t_i,t_j)$ refers to the cosine similarity\footnote{We employ the NLTK toolkit to compute the semantic similarity of a pair of tokens. The NLTK toolkit can be accessed via~\url{http://www.nltk.org}.}, representing the semantic similarity of tokens $t_i$ and $t_j$. \textcolor{black}{The rationale for adopting cosine similarity is that it is a prevalent metric for effectively assessing the semantic similarity between two tokens~\cite{liang-etal-2022-multi,hu-etal-2021-mmgcn}.}
$\lvert p_i-p_j\rvert$ is used to measure the sentiment similarity. \textcolor{black}{$p_i$ and $p_j$ are the polarity of $t_i$ and $t_j$, respectively.} 
$w(n_i,n_j)$ refers to the weight of the edges constructed for representing sentiment-oriented semantic relation between the nodes $n_i$ and $n_j$. To normalize the weight of these edges, we set its maximum value as $1$. 

Accordingly, the adjacency matrix $\mathbf{A}^2 \in \mathbb{R}^{N \times N}$ for capturing the above sentiment-oriented semantic relations can be constructed as follows, 
\begin{equation}
 \mathbf{A}^2_{i,j}=\left\{
 \begin{aligned}
w(n_i,n_j) & ,\quad if~D_2(n_i,n_j), \\\
0 & ,\quad otherwise,
\end{aligned}\label{eq:graph_1}
 \right.
\end{equation}
where $D_2(n_i, n_j)$ indicates that nodes $n_i$ and $n_j$ have certain above sentiment-oriented semantic relation, $i\in [1, N_t]$ and $j\in [N_t+1, N]$. $N$ is the total number of nodes in the graph. 

Ultimately, by combing the adjacency matrices for context-oriented and sentiment-oriented semantic relations, \ie $\mathbf{A}^1$ and $\mathbf{A}^2$, we can derive the final adjacency matrix $\mathbf{A}$ for the context-sentiment graph. \textcolor{black}{We illustrate the context-sentiment graph construction for the given dialogue in Fig.~\ref{fig:graph}.} 

\subsubsection{Graph Convolution Network}
Towards the final context encoding, we adopt $L$ layers of GCN. Then the node representations are iteratively updated as follows,
\begin{equation}
    \mathbf{G}_{l}=ReLU(\tilde{\mathbf{A}}\mathbf{G}_{l-1}\mathbf{W}_l), l \in [1,L],
\end{equation}
where $\tilde{\mathbf{A}} = (\mathbf{D})^{-\frac{1}{2}}\mathbf{A}(\mathbf{D})^{-\frac{1}{2}}$ is the normalized symmetric adjacency matrix, and $\mathbf{D}$ is the degree matrix of the adjacency matrix $\mathbf{A}$. $\mathbf{W}_l \in \mathbb{R}^{D \times D}$ are trainable parameters of the $l$-th GCN layer. 
$\mathbf{G}_l$ are the node representations obtained by the $l$-th layer, where 
$\mathbf{G}_0=\mathbf{H}$ is the initial node representation.

\subsection{Sarcasm Explanation Generation}
The final nodes representation $\mathbf{G}_L$ obtained by the $L$-th layer GCNs absorb rich semantic information from their correlated nodes and can be used as the input for the following sarcasm explanation generation. Considering the promising performance of residual connection in the task of 
text generation~\cite{DBLP:conf/nips/VaswaniSPUJGKP17,DBLP:conf/acl/JingSOJN23}, we also introduce a residual connection for generating the sarcasm explanation as follows,
\begin{equation}
\mathbf{R}=\mathbf{H}+\mathbf{G}_L,
\end{equation}
where $\mathbf{R} \in \mathbb{R}^{N\times D}$ denotes the fused node representation.
We then feed $\mathbf{R}$ to the decoder of the pre-trained BART. The decoder works in an auto-regressive manner, namely, producing the next token by considering all the previously decoded outputs as follows,
\begin{equation}
\hat{\textbf{y}}_t=Decoder_B(\mathbf{R}, \hat{Y}_{<t}),
\end{equation}
where $t \in [1,N_y]$ and $\hat{\textbf{y}}_t \in \mathbb{R}^{|\mathcal{V}|}$ is the predicted $t$-th token's probability distribution of the target sarcasm explanation, $Decoder_B$ refers to the BART decoder. $\hat{Y}_{<t}$ refers to the previously predicted $t$-$1$ tokens.

%\textbf{Training.}
For optimization, we adopt the cross-entropy loss as follows,
\begin{equation} \label{prob}
  \mathcal{L} = -1/N_y\sum_{i=1}^{N_y}\log (\hat{\textbf{y}}_i[t]),
 \end{equation}
where $\hat{\textbf{y}}_i[t]$ is the element of  $\hat{\textbf{y}}_i$ that corresponds to the $i$-th token of the target explanation, and $N_y$ is the total number of tokens in the target sarcasm explanation $Y$.

\section{Experiments}
\subsection{Experimental Settings}

\textbf{Dataset.}
In this work, we adopted the public dataset named WITS~\cite{DBLP:conf/acl/KumarKA022} for SED task. It is a multimodal, multi-party, Hindi-English-mixed dialogue dataset collected from the popular Indian TV show, `Sarabhai v/s Sarabhai'\footnote{\url{https://www.imdb.com/title/tt1518542/}}. And it consists of $2,240$ sarcastic dialogues. Each dialogue is associated with the corresponding utterances, video, audio, and manual annotated sarcasm explanation.
The number of utterances ranges from $2$ to $27$ for dialogues. 
We adopted the original setting~\cite{DBLP:conf/acl/KumarKA022}, the ratio of data split for training/validation/testing sets is $8:1:1$ 
for experiments, resulting in $1,792$ dialogues in the training set and $224$ dialogues each in the validation and testing sets.

\begin{figure}[!t]
\centering
 \includegraphics[scale=0.58]{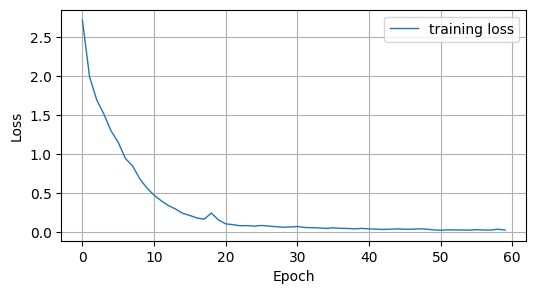}   
   \caption{\textcolor{black}{The training curve for our EDGE in $60$ epochs.}}
   \label{fig:loss}
\end{figure}

\textbf{Implementation Details.}
To verify the effectiveness of our method in different backbones, following the backbone settings of $\textbf{MAF-TAV}_B$ and $\textbf{MAF-TAV}_M$~\cite{DBLP:conf/acl/KumarKA022}, we also adopt BART-base\footnote{\url{https://huggingface.co/facebook/bart-base}.} and mbart-large-50-many-to-many-mmt\footnote{\url{https://huggingface.co/facebook/mbart-large-50-many-to-many-mmt}.} as the backbone of our model, respectively. 
\textcolor{black}{Following the original setting~\cite{DBLP:conf/aaai/KumarMA023}, the total number of tokens for the input text,} \ie $N$, is unified to $480$ by padding or truncation operations. 
The feature dimension $d_a$, $d_v$, $d$ and $D$ of the audio, video, concatenated feature $\mathbf{J}$ and the encoded representation matrix $\mathbf{H}$ are set to $512$, $512$, $1024$ and $768$, respectively.
In addition, the resampled clip size $N_c$ of the video and audio clips is fixed to $8$. We used AdamW~\cite{DBLP:journals/corr/abs-1711-05101} as the optimizer and set the learning rate of GCNs to $10e$-$4$ and that of the BART to $5e$-$5$. 
The batch size is set to $16$ and the maximum number of epochs for model training is set to $60$. 
\textcolor{black}{Fig.~\ref{fig:loss} visualizes the training process, where the training loss steadily decreases with minor fluctuations until the best performance is achieved.} 
Following the previous work~\cite{DBLP:conf/acl/KumarKA022}, we employed \mbox{BLEU-1}, \mbox{BLEU-2}, \mbox{BLEU-3}, \mbox{BLEU-4}~\cite{DBLP:conf/acl/PapineniRWZ02}, \mbox{ROUGE-1}, \mbox{ROUGE-2}, \mbox{ROUGE-L}~\cite{Lin2004ROUGEAP}, METEOR~\cite{DBLP:conf/coling/BabanejadDAP20}, BERT-Score~\cite{DBLP:conf/iclr/ZhangKWWA20} to evaluate the performance of sarcasm explanation generation models. For all the metrics, the larger the better.

\begin{table*}[h]
\centering
\caption{\textcolor{black}{Performance (\%) comparison among different methods on WITS. The best results are in boldface, while the second best are underlined. $\star$ denotes that the p-value of the significance test between our result and the best baseline MOSES result is less than 0.01. ''Improvement~$\uparrow$``: the relative improvement by our model over the best baseline.}  
}
\label{tab:model_comp}
\resizebox{\textwidth}{!}{%
\begin{tabular}{c|ccccccccc}
\hline
{ \textbf{Model}} & \textbf{ROUGE-1} & \textbf{ROUGE-2} & \textbf{ROUGE-L} & \textbf{BLEU-1} & \textbf{BLEU-2} & \textbf{BLEU-3} & \textbf{BLEU-4} & \textbf{METEOR} & \textbf{BERT-Score} \\ \hline
\textbf{RNN}~\cite{DBLP:conf/acl/KleinKDSR17}~(2017)                          & 29.22           & 7.85            & 27.59           & 22.06          & 8.22           & 4.76           & 2.88           & 18.45            & 73.24              \\
\textbf{Transformers}~\cite{DBLP:conf/nips/VaswaniSPUJGKP17}~(2017)                 & 29.17           & 6.35            & 27.97           & 17.79          & 5.63           & 2.61           & 0.88           & 15.65            & 72.21              \\
\textbf{PGN}~\cite{DBLP:conf/acl/SeeLM17}~(2017)                          & 23.37           & 4.83            & 17.46           & 17.32          & 6.68           & 1.58           & 0.52           & 23.54            & 71.90              \\
\textbf{BART}~\cite{DBLP:conf/acl/LewisLGGMLSZ20}~(2020)                          & 36.88           & 11.91           & 33.49           & 27.44          & 12.23          & 5.96           & 2.89           & 26.65            & 76.03              \\
\textbf{mBART}~\cite{DBLP:journals/tacl/LiuGGLEGLZ20}~(2020)                        & 33.66           & 11.02           & 31.50           & 22.92          & 10.56          & 6.07           & 3.39           & 21.03            & 73.83              \\
 \hline

$\textbf{MAF-TAV}_M$~\cite{DBLP:conf/acl/KumarKA022}~(2022)                  & 38.52           & 14.13           & 36.60           & 30.50          & 15.20          & 9.78          & 5.74           & 27.42            & 76.70              \\ 

$\textbf{MAF-TAV}_B$~\cite{DBLP:conf/acl/KumarKA022}~(2022)                      & 39.69           & 17.10           & 37.37           & 33.20          & 18.69          & 12.37          & 8.58           & 30.40            & 77.67              \\

\textcolor{black}{$\textbf{Video-LLaMA}$~\cite{damonlpsg2023videollama}~(2023)}                      & 39.74           & 17.95           & 37.56           & 31.93          & 19.31          & 13.07          & 8.92           & 30.92            &76.89              \\
\textcolor{black}{$\textbf{Video-ChatGPT}$~\cite{Maaz2023VideoChatGPT}~(2024)}                      & 41.02           & 19.72          & 38.91           & 32.71          & 20.53          & 14.59          & 10.54           & 31.67            & 77.8              \\
$\textbf{MOSES}$~\cite{DBLP:conf/aaai/KumarMA023}~(2022)                      & 42.17           & 20.38           & 39.66           & \underline{34.95}          & 21.47          & \underline{15.47}          & \underline{11.45}           & 32.37            & 77.84              \\
\hline

$\textbf{EDGE}_M$                            & \underline{43.74}           &  \underline{20.80}           &  \underline{39.98}           &  34.91          &  \underline{21.56}          &  14.06          &  10.19          &  \underline{37.52}              &  \underline{78.81}                 \\
$\textbf{EDGE}$                            & $\textbf{44.35}^{\star}$          & $\textbf{21.76}^{\star}$         & $\textbf{42.38}^{\star}$          & $\textbf{37.64}^{\star}$      & $\textbf{23.23}^{\star}$        & $\textbf{16.58}^{\star}$         & $\textbf{12.85}^{\star}$         & $\textbf{39.88}^{\star}$             & $\textbf{80.21}^{\star}$                \\ \hline
\textbf{Improvement}~$\uparrow$                      & 2.18           & 1.38           & 2.72           & 2.69          & 1.76          & 1.11          & 1.40           & 7.51            & 2.37              \\ \hline
\end{tabular}%
}
\end{table*}
\subsection{On Model Comparison}
For evaluation,  we compared our EDGE with the following baselines, including text-based models (\ie \textbf{RNN}, \textbf{Transformers}, \textbf{PGN}, \textbf{BART} and \textbf{mBART}) and multimodal models (\ie $\textbf{MAF-TAV}_M$, $\textbf{MAF-TAV}_B$, \textcolor{black}{\textbf{Video-LLaMA}, \textbf{Video-ChatGPT}}, \textbf{MOSES}, and $\textbf{EDGE}_M$).  
\begin{itemize}
    \item \textbf{RNN}~\cite{DBLP:conf/acl/KleinKDSR17}. This is a classical seq-to-seq architecture, which can process sequential data and is easy to extend. 
    The openNMT\footnote{\url{https://github.com/OpenNMT/OpenNMT-py}.} implementation of the RNN seq-to-seq architecture is used in our experiment.
    \item \textbf{Transformers}~\cite{DBLP:conf/nips/VaswaniSPUJGKP17}. This text-based generation baseline generates the explanation with the advanced Transformer.
    \item \textbf{PGN}~\cite{DBLP:conf/acl/SeeLM17}. Pointer Generator Network is a text-based generation model, which generates the text with not only a conventional decoder but also a copy mechanism that copies words directly from input text. 
    \item \textbf{BART}~\cite{DBLP:conf/acl/LewisLGGMLSZ20}. It is a denoising auto-encoder model with standard Transformer architecture, and pretrained for natural language generation, translation, and comprehension. 
    \item \textbf{mBART}~\cite{DBLP:journals/tacl/LiuGGLEGLZ20}. It has the same architecture as BART and is pretrained on a large-scale multilingual corpus. 
     \item  $\textbf{MAF-TAV}_M$ and $\textbf{MAF-TAV}_B$~\cite{DBLP:conf/acl/KumarKA022}.  
     To use the multimodality information, they employ mBART and BART as the backbone, respectively, where a  modality-aware fusion module is devised to fuse multimodal information. 
\item \textbf{Video-LLaMA}~\cite{damonlpsg2023videollama}. It integrates the visual encoder BLIP-2~\cite{DBLP:conf/icml/0008LSH23}, audio encoder ImageBind~\cite{DBLP:conf/cvpr/GirdharELSAJM23}, and the large language model LLaMA~\cite{DBLP:journals/corr/abs-2302-13971}, to perform spatial-temporal modeling for videos.
\item \textbf{Video-ChatGPT}~\cite{Maaz2023VideoChatGPT}. This is an adapted multimodal large language model~\cite{DBLP:conf/nips/LiuLWL23a}, integrated with the visual encoder CLIP~\cite{DBLP:conf/icml/RadfordKHRGASAM21} and the language decoder Vicuna~\cite{vicuna2023}, which can perform spatial-temporal video representation.
\item \textbf{MOSES}~\cite{DBLP:conf/aaai/KumarMA023}. To incorporate the multimodal information, it adopts BART as the backbone, where a multimodal context-aware attention module is devised to fuse multimodal information. 
\item  $\textbf{EDGE}_M$. The model is a variant of EDGE in which mBART is adopted as the backbone instead of BART.  
\end{itemize}

\textbf{Objective Evaluation.} Table~\ref{tab:model_comp} shows the performance comparison among different methods, where we also conduct the significance test. Specifically, we train both EDGE and the best baseline MOSES ten times, each with a different random seed. We then conduct t-test~\cite{ref1} to calculate the P-value for each metric. From this table, we have the following several observations. 
1) Our model EDGE exceeds all the baselines in terms of all the metrics, and our variant model $\text{EDGE}_M$ with mBART as backbone also outperforms baselines on most evaluation metrics. This comprehensively demonstrates the superiority of our model in SED. 
2) EDGE outperforms the $\text{EDGE}_M$, which is consistent with the observation that 
\textcolor{black}{BART has a better performance than mBART. In fact, among all the text-based models, BART performs best, which shows the strong generation capability of BART in the context of SED.  
The reasons can be two folds. On the one hand, though the input utterances are Hindi-English mixed, the Romanized Hindi in the dataset closely aligns with English, which facilitates the fine-tuning of BART for understanding the Hindi part of the input~\cite{DBLP:conf/acl/KumarKA022}. On the other hand, mBART is pre-trained for multilingual tasks on a wide range of languages, while our study concentrates on Romanized Hindi and English. Then the multilingual capabilities of mBART, while robust, may introduce unnecessary noise due to the inclusion of languages beyond our scope of interest. 3) Multimodal models (\ie $\text{MAF-TAV}_M$, $\text{MAF-TAV}_B$, \text{Video-LLaMA}, \text{Video-ChatGPT}, \text{MOSES} and $\text{EDGE}_M$) have a better performance than text-based models (\ie \text{RNN}, \text{Transformers}, \text{PGN}, \text{BART} and \text{mBART}), which verifies that the video and audio modalities can provide useful information for the sarcasm explanation generation.} 
4) Unexpected, Video-LLaMA, which can leverage all the video, audio and text inputs for SED, underperforms Video-ChatGPT that is limited to only video and text inputs. 
The underperformance may stem from the fact that compared to the pooling mechanism employed in Video-ChatGPT, the Q-former~\cite{DBLP:conf/icml/0008LSH23} used in Video-LLaMA compresses the number of visual tokens by abstracting semantic-level visual concepts, leading to visual semantics deficiency (\eg the loss of fine-grained attributes and spatial locality~\cite{Yao2024DeCoDT}), and causing the degradation of video comprehension capacity~\cite{DBLP:conf/mm/XuZX0Z0Z17,DBLP:conf/aaai/YuXYYZZT19}. 
5) Multimodal large language models (\ie \text{Video-LLaMA} and \text{Video-ChatGPT}) underperform our EDGE, it further proves the advantage of utilizing sentiments to enhance sarcasm semantics comprehension, since Video-LLaMA and Video-ChatGPT overlook the sentiments in the multimodal input.

\textbf{Human Evaluation.} To thoroughly assess the quality of generated explanations and verify the superiority of EDGE, we also conduct human evaluation between our EDGE and the best baseline MOSES. Given that the WITS dataset provides both the original multilingual dialogue data for model processing and its English translations, where Hindi utterances are translated into English for human understanding, we employ three volunteers proficient in English to perform human evaluation.
Each volunteer needs to evaluate $224$ dialogue samples. For each sample, the volunteers are required to select the more plausible explanation from a pair of explanations from our EDGE and MOSES according to the following three aspects.
\begin{itemize}
    \item  \textbf{Fluency}: whether the explanation is expressed fluently. 
    \item  \textbf{Relevance}: whether the explanation revolves around the topic of the dialogue. 
    \item \textbf{Validity}: whether the explanation captures the sarcasm in the dialogue. 
\end{itemize}
\begin{table}[!t]
\setlength{\abovecaptionskip}{0cm}
\setlength{\belowcaptionskip}{0cm}
\caption{\textcolor{black}{Human evaluation for explanations generated by EDGE and the best baseline MOSES.}}
\centering
\setlength{\tabcolsep}{1mm}{
\begin{tabular}{c|ccc}
\hline
\textbf{Evaluation Factors} & \textbf{Wins (\%)} & \textbf{G-$\gamma$ (\%)} & \textbf{C-$\kappa$ (\%)} \\ \hline
\textbf{Fluency}  & 63.8   & 79.6 & 72.5 \\
\textbf{Relevance}  & 66.5   & 75.4 & 69.7\\
\textbf{Validity}  & 69.2  & 71.2 & 65.9\\ \hline  
\end{tabular}
}
\label{Tab:human_eva}
\vspace{-0.8em}
\end{table}
In the evaluation process, the volunteers do not know the explanation is generated by which model,
and the final verdict for each pair is determined by a majority vote among the three volunteers. Table~\ref{Tab:human_eva} shows the human evaluation results and the inter-annotator agreement with respect to both Gwet's $\gamma$ \cite{Gwet2014HandbookOI} and Cohen's $\kappa$~\cite{Cohen1960ACO}. As we can see, our EDGE wins MOSES on more than $60.0$\% samples across all the three evaluation aspects, which further demonstrates the superiority of our EDGE. Across all three aspects, Gwet's $\gamma$ values exceed $70.0\%$ and Cohen's $\kappa$ values surpass $60.0\%$, which mean substantial agreement.
It statistically verifies the inter-annotator consistency and reliability of the human evaluation.

\textcolor{black}{\textbf{Complexity and Efficiency Comparison.} To learn the complexity and efficiency of our model, we show the number of parameters and the inference speed of our model and all multimodal baselines in Table~\ref{Tab:complexity&efficiency}. To ensure a fair comparison, all model inference processes are conducted on a single A800 80GB GPU with a maximum of 256 CPU cores.}
As we can see, compared with BART-based baselines (\ie $\text{MAF-TAV}_B$, $\text{MOSES}$), our EDGE offers a simpler framework with fewer parameters. Meanwhile, our $\text{EDGE}_M$ also involves fewer parameters than $\text{MAF-TAV}_M$, both of which are based on mBART. As expected, the two multimodal large language models, \ie Video-LLaMA and Video-ChatGPT, involve significantly more parameters. 
In addition, the efficiency of our \text{EDGE} exceeds all the multimodal baselines, and $\text{EDGE}_M$ is comparable to the two most efficient baselines \ie $\text{MAF-TAV}_B$ and $\text{MAF-TAV}_M$. 
Notably, \text{Video-LLaMA} and \text{Video-ChatGPT} exhibit diminished efficiency due to their complex framework.

\begin{table}[!t]
\setlength{\abovecaptionskip}{0cm}
\setlength{\belowcaptionskip}{0cm}
\caption{\textcolor{black}{Complexity and efficiency comparison results. 
\textbf{Time} is the average time consumption of samples in the testing set.}}
\label{Tab:complexity&efficiency}
\centering
\setlength{\tabcolsep}{1mm}{
\begin{tabular}{c|c|c|c}
\hline
\textbf{Model} &\textbf{Backbone} &\textbf{\#Params} & \textbf{Time}      \\ \hline 
{$\textbf{MAF-TAV}_M$}&mBART& 1147M & 1.4s        \\
{$\textbf{MAF-TAV}_B$} & BART & 177M & 1.2s       \\
{$\textbf{Video-LLaMA}$}&LLaMA+ImageBind& 7B& 3.5s\\
{$\textbf{Video-ChatGPT}$}&Vicuna+CLIP& 7B& 3.8s \\
{$\textbf{MOSES}$}& BART& 326M& 1.7s\\ \hline
{$\textbf{EDGE}_M$}&mBART & 1124M& 1.5s\\
{$\textbf{EDGE}$} &BART &154M& 1.1s\\
\hline
\end{tabular}
}
\vspace{-0.5em}
\end{table}

\begin{table*}[t]
\caption{Ablation study results (\%) of our proposed EDGE. The best results are highlighted in boldface.}
\label{tab:ablation}
\resizebox{\textwidth}{!}{
\begin{tabular}{c|ccccccccc}
\hline
{ \textbf{Model}} & \textbf{ROUGE-1} & \textbf{ROUGE-2} & \textbf{ROUGE-L} & \textbf{BLEU-1} & \textbf{BLEU-2} & \textbf{BLEU-3} & \textbf{BLEU-4} & \textbf{METEOR} & \textbf{BERT-Score} \\ \hline
\textbf{w/o-U-Content}                      & 27.01           & 6.49            & 25.18           & 21.77          & 7.33           & 2.73           & 1.65           & 25.20             & 71.10               \\
\textbf{w/o-U-Sentiment}                         & 43.67           & 21.19           & 40.02           & 35.86          & 22.60          & 16.29          & 12.09          & 35.64             & 78.46               \\ \hline
\textbf{w/o-VA-Sentiment}                         & 43.33           & 20.32           & 40.75           & 35.64          & 21.80          & 14.90          & 10.20          & 37.81          &79.50                   \\
\textbf{w-VA-Content}                        & 39.74           & 16.92           & 37.52           & 32.13          & 17.32          & 11.26          & 8.64          & 32.11             & 75.51               \\ 
 \hline
 \textbf{w/o-GCNs}                    & 41.34            & 18.75            & 38.74            & 33.46           & 19.90           & 13.83           & 9.79           & 36.23             & 77.49               \\
 \textbf{w/o-U-Relation}
& 43.18           & 20.26           & 41.84           & 34.26          & 21.89          & 15.64          & 11.92          & 37.41             & 76.51  
\\
\textbf{w/o-S-Relation}
& 43.07           & 20.79           & 41.19           & 34.76          & 22.21          & 15.87          & 11.68          & 37.29             & 78.34               \\ 
\textcolor{black}{\textbf{w/o-SentimentNode}} &42.72 &20.17 &39.95 &34.91 &21.13 &14.50 &9.93 &35.39 &77.95
\\
\hline
\textbf{w/o-Weight}                    & 43.42           & 21.57           & 41.31           & 35.62          & 22.53          & 16.26          & 12.55          & 37.98             & 78.21               \\
\textcolor{black}{\textbf{w-ED-Weight}}&   43.11         &  21.08         & 41.67       & 35.72          &21.43         & 16.62          & 11.86       & 39.10          &  79.14 
\\
\textcolor{black}{\textbf{w-MMD-Weight}}&   42.62         &  20.26         & 40.13       & 34.10          &20.23         & 15.11          & 10.71       & 37.05          &  77.37 
\\
\textcolor{black}{\textbf{w-CMD-Weight}}&   42.79         &  20.98         & 40.77       & 35.02          &20.93         & 16.25          & 11.02       & 38.17          &  78.29 
\\
\textcolor{black}{\textbf{w-LearnableWeight}} &   42.81         &  20.54         & 41.51       & 35.17          &21.65         & 15.96          & 10.94       & 38.27          &  78.39 
\\

\hline
$\mathbf{EDGE}$                          & \textbf{44.35}           & \textbf{21.76}           & \textbf{42.38}           & \textbf{37.64}          & \textbf{23.23}          & \textbf{16.58}          & \textbf{12.85}          &\textbf{39.88}              & \textbf{80.21}               \\ \hline
\end{tabular}%
}
\end{table*}

\begin{figure*}[!t] 
    \centering
    \includegraphics[scale=0.495]{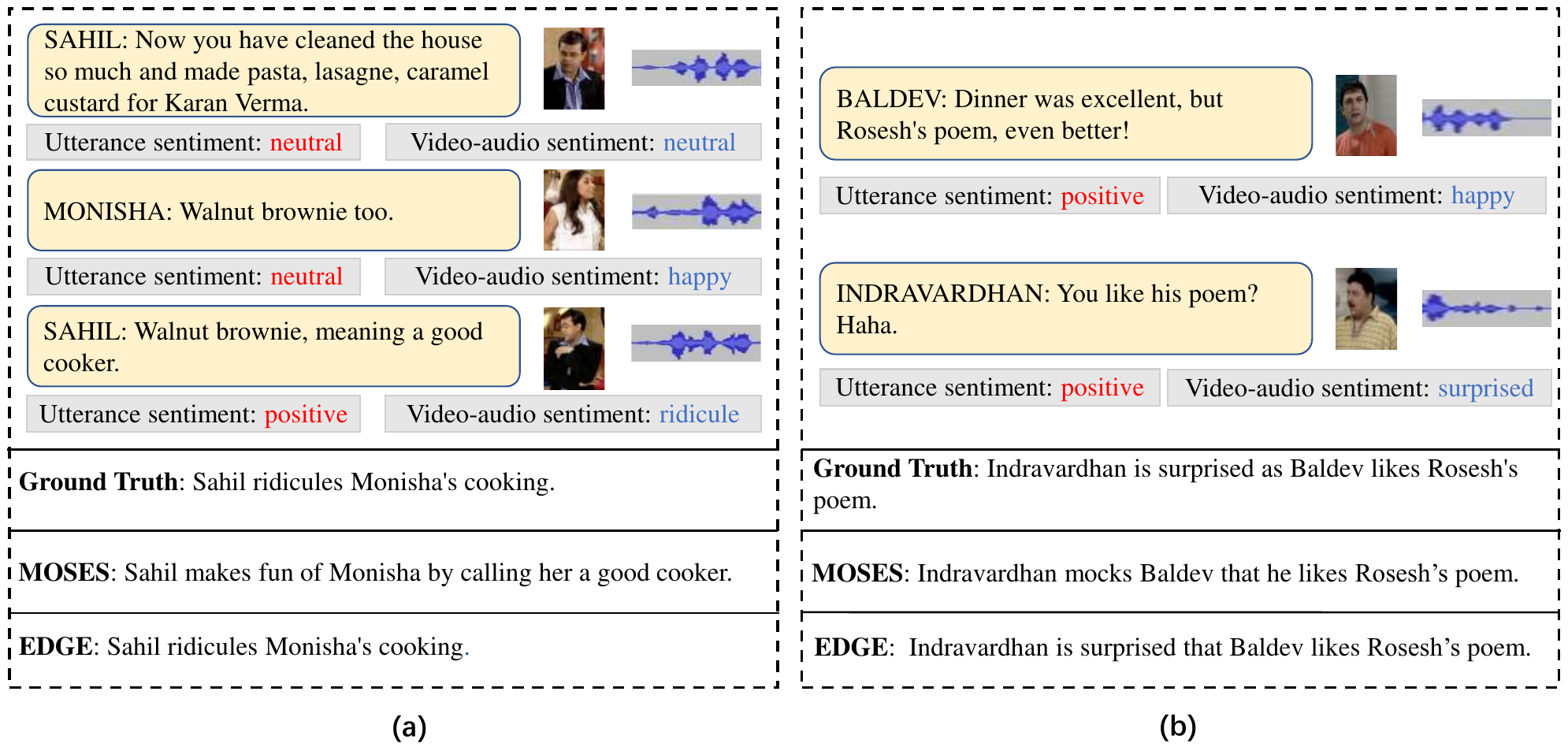}
    \caption{Comparison between the explanation generated by our EDGE and the best baseline $\textbf{MOSES}$ on two testing samples.}
    \label{fig:case}
\end{figure*}

\subsection{On Ablation Study}
We introduced various variants of our model in order to explore the contribution of each component in EDGE. 

For the lexicon-guided utterance sentiment inference module, we devised the following two variants of EDGE. 1) \textbf{w/o-U-Content}. To evaluate the role of the utterances in the dialogue, we did not utilize the utterances content in this variant. 2)  \textbf{w/o-U-Sentiment}. To show the importance of the sentiments inferred from the utterances, we omitted the lexicon-guided utterance sentiment inference module. 

For the video-audio joint sentiment inference module, we introduced two variants of EDGE. 1) \textbf{w/o-VA-Sentiment}. To show the benefit of the video-audio sentiments, we removed the video-audio joint sentiment inference module. 2) \textbf{w-VA-Content}. To demonstrate the advantages of utilizing the video-audio sentiments over the direct input of video and audio modality information, we concatenated visual and acoustic features with textual features to derive the encoded representation matrix $\mathbf{H}$ instead of using the video-audio sentiments.

For the sentiment-enhanced context encoding module, we designed the following variants of EDGE. 1) \textbf{w/o-GCNs}.
To verify the necessity of modeling the semantic relations with GCNs, we removed the context-sentiment graph and GCNs. Specifically, we directly fed the encoded representation matrix $\mathbf{H}$ into the BART decoder.
2) \textbf{w/o-U-Relation}.
To prove the validity of the context-oriented semantic relation in the context-sentiment graph, we removed the context-oriented semantic relation.
3) \textbf{w/o-S-Relation}.
To verify the effectiveness of the sentiment-oriented semantic relation in the context-sentiment graph, we omitted the sentiment-oriented semantic relation.
4) \textcolor{black}{\textbf{w/o-SentimentNode}. To explore the role of sentiments in context-sentiment graph, we removed both utterance sentiment nodes and video-audio sentiment nodes from the graph.}
5) \textbf{w/o-Weight}. To show the effectiveness of our defined weights for sentiment-oriented semantic relations, we replaced all the weights of these edges (\ie utterance sentiment-content edges, video-audio sentiment-content edges, and sentiment-sentiment edges) with $1$.
6) \textbf{w-ED-Weight}, \textbf{w-MMD-Weight}, and \textbf{w-CMD-Weight}. To demonstrate the superiority of using cosine similarity in the weight calculation for sentiment-oriented semantic relations, we replaced cosine similarity with Euclidean Distance (ED), Maximum Mean Discrepancy (MMD), and Central Moment Discrepancy (CMD), respectively.
7) \textbf{w-LearnableWeight}. In this variant, we replaced GCNs by Graph Attention Networks (GAT) to learn the weights of edges automatically.

The ablation study results are shown in Table~\ref{tab:ablation}. From this table, we have the following observations. 
1) EDGE outperforms w/o-U-Content and w/o-U-Sentiment, which verifies that both utterance content and utterance sentiments are helpful in understanding the ironic semantics.
2) EDGE performs better than w/o-VA-Sentiment and w-VA-Content. It demonstrates that video-audio sentiments do assist sarcastic semantic comprehension, and proves the superiority of utilizing the video-audio sentiments compared with directly inputting the visual and acoustic features. 
3) \textcolor{black}{EDGE performs better than w/o-GCNs, w/o-U-Relation, w/o-S-Relation, and w/o-SentimentNode. It verifies the superiority of modeling the given dialogue by GCNs with our proposed context-sentiment graph. Meanwhile, it shows the effectiveness of context-oriented semantic relations, sentiment-oriented semantic relations, and sentiment nodes in capturing the sarcastic semantics.}
4) \textcolor{black}{EDGE consistently exceeds w/o-Weight, w-ED-Weight, w-MMD-Weight, w-CMD-Weight, and w-LearnableWeight. This proves the advantage of our proposed cosine similarity-based weighting strategy for sentiment-oriented semantic relations in the context-sentiment graph. Meanwhile, it reflects that although GAT can learn weights automatically, it may struggle to capture the complex semantic relations with limited training data in the context of SED.}

\begin{figure*}[h]
    \centering
    \includegraphics[scale=0.495]{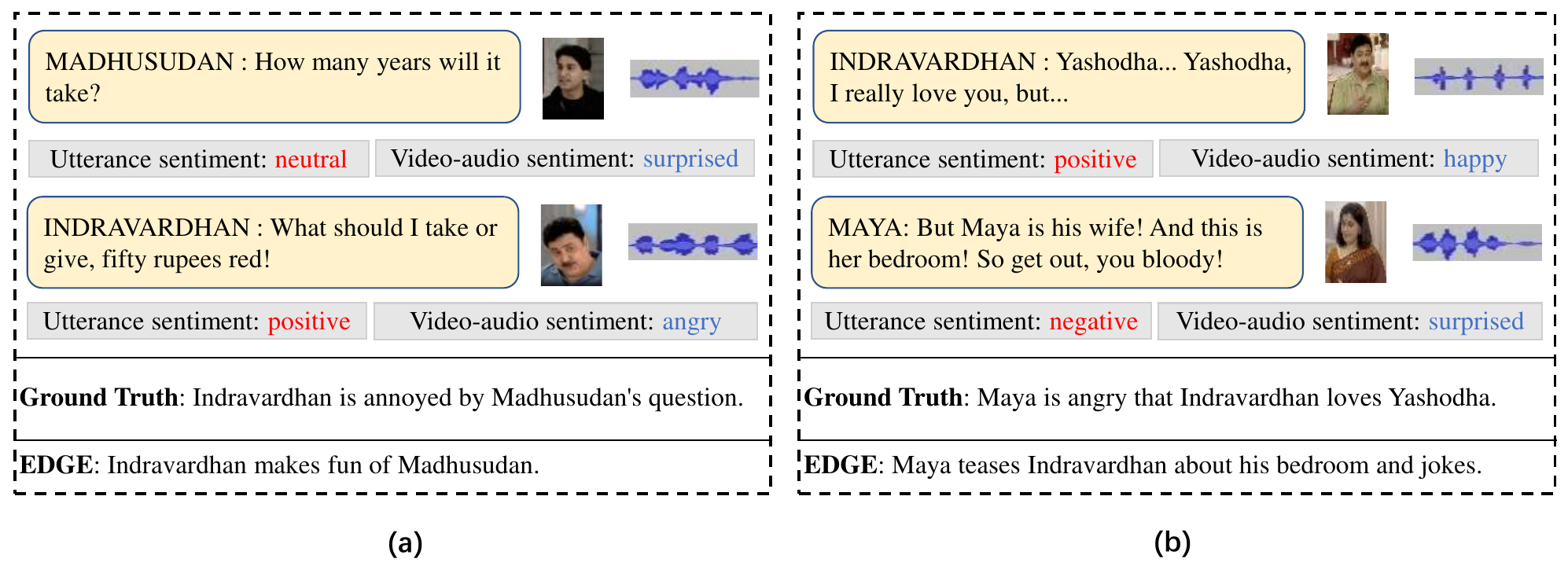}
       % \vspace{-0.5cm}
    \caption{The error cases where our EDGE failed to generate an appropriate explanation.}
     % \vspace{-0.4cm}
    \label{fig:error_cases}
\end{figure*}
\subsection{On Case Study}
To get an intuitive understanding of how our model works on Sarcasm Explanation in Dialogue, we first show two testing samples in Fig.~\ref{fig:case}. For comparison, we also displayed the sarcasm explanation generated by the best baseline MOSES.
In case (a), our model performs better than MOSES in terms of the quality of the generated sarcasm explanation, as the sarcasm explanation generated by our EDGE is the same as the ground truth. It is reasonable since the video-audio sentiment ``ridicule'' inferred in the last utterance boosts the sarcasm explanation generation. In addition, for the last utterance, the utterance sentiment ``positive'' and the video-audio sentiment ``ridicule'' are obviously inconsistent, which may provide vital clues for sarcastic semantic comprehension and explanation generation. In case (b), our model properly explains the sarcasm involved in the dialogue, while MOSES failed. By analyzing the extracted video-audio sentiments, we noticed that the video-audio sentiment ``surprised'' benefits the semantics comprehension of the input dialogue and hence promote the sarcasm explanation generation. Overall, these two cases intuitively show the benefits of incorporating both utterance sentiments and video-audio sentiments into the context of sarcasm explanation in dialogue.

\textcolor{black}{Moreover, we also exhibit two error cases of our EDGE in Fig.~\ref{fig:error_cases}. As can be seen, in case (a), the phrase ``fifty rupees red'' is a colloquial or idiomatic expression in Hindi, which likely confuses EDGE due to its lack of exposure to such cultural nuances. In case (b), ``Yashodha'' refers to a character from Indian mythology, which further challenges EDGE to fully understand the context. These examples highlight the need for external knowledge to effectively capture sarcasm in culturally specific cases, indicating a potential avenue for further improving the performance of SED.}

\section{Conclusion and Future Work}
In this work, we propose a novel sentiment-enhanced Graph-based multimodal sarcasm
Explanation framework named EDGE, which incorporates the utterance sentiments and video-audio sentiments into the context of the dialogue to improve sarcasm explanation in dialogue. The experiment results on WITS dataset demonstrate the superiority of our model over the existing cutting-edge methods, and validate the benefits of the utterance sentiments, video-audio sentiments, as well as the context-sentiment graph, which can fully model the semantic relations among the utterances, utterance sentiments, and video-audio sentiments, including context-oriented semantic relation and sentiment-oriented semantic relation. In the future, we plan to adopt more advanced large language models such as GPT-4o to improve SED task.

\bibliographystyle{IEEEtran}
\balance
\bibliography{reference}

\vfill

\end{document}